\documentclass[preprint,12pt]{elsarticle}

\usepackage{amssymb}
\usepackage{amsmath}
\journal{}

\begin{document}

\begin{frontmatter}



\title{Navigating Decision Landscapes: The Impact of Principals on Decision-Making Dynamics}


\author[inst1]{Lu Li}

\affiliation[inst1]{organization={Microsoft},%Department and Organization
            addressline={One Microsoft Way}, 
            city={Redmond},
            postcode={98052}, 
            state={WA},
            country={US}}

\author[inst1]{Huangxing Li}

\begin{abstract}

We explored decision-making dynamics in social systems, referencing the 'herd behavior' from prior studies where individuals follow preceding choices without understanding the underlying reasons. While previous research highlighted a preference for the optimal choice without external influences, our study introduced principals or external guides, adding complexity to the decision-making process. The reliability of these principals significantly influenced decisions. Notably, even occasional trust in an unreliable principal could alter decision outcomes. Furthermore, when a principal's advice was purely random, heightened trust led to more decision errors. Our findings emphasize the need for caution when placing trust in decision-making contexts.
\end{abstract}


\end{frontmatter}


\section{Introduction}
\label{sec:sample1}

%% Sequential decision-making is important
Sequential decision-making is a cornerstone of cognitive science, economics, and information theory. It encapsulates the intricate process by which individuals navigate multifaceted environments. In today's digital age, where we are constantly bombarded with vast amounts of data and information from various sources, decisions are seldom made in a vacuum. Instead, they often result from a series of preceding choices, both personal and those observed from others.

A pivotal element in this decision-making process is the role of principal signals. These are vital cues or pieces of information that steer the decision-making trajectory. For instance, consider the financial market. Here, a principal could be a renowned financial analyst or institution whose forecasts and recommendations significantly influence investor behavior. When this analyst predicts a bullish trend for a particular stock, many investors, especially those who trust or follow this analyst, might decide to buy the stock based on this principal signal, even if they haven't done their own thorough research. Such principal signals, whether they originate from expert opinions, media reports, or influential figures, play a crucial role in shaping public opinion and actions.

The ability to discern these signals, coupled with insights from prior decisions, paves the way for more strategic and informed choices. Thus, the significance of sequential decision-making is profound, offering insights into the complex interplay of cognition, information, and social influences in shaping individual choices.

In this study, we embark on an exhaustive journey, addressing several pivotal research questions that illuminate the intricacies of decision-making under diverse scenarios. Our initial exploration probes the influence of principals on decision convergence. Do certain guiding signals from principals lead to more harmonized decisions? We then investigate how varied principals' settings, their reliability, and the trust placed in them sculpt the decision-making path.

Our research findings, derived from extensive simulations, reveal a nuanced landscape of decision-making influenced by principal signals. We observed that the presence of a reliable principal can expedite convergence towards optimal decisions. However, when the principal's guidance is inconsistent or misleading, it can introduce significant variability into the decision-making process, sometimes leading to suboptimal outcomes. Furthermore, the degree of trust vested in the principal emerged as a critical factor. Even occasional trust in a predominantly unreliable principal can profoundly alter decision trajectories, underscoring the importance of discernment in decision-making trust allocation. Conversely, when the principal's guidance was statistically random, an intriguing inverse relationship between trust in the principal and decision accuracy was noted.

These findings have broader implications for various sectors, from finance to public policy, emphasizing the need for critical evaluation of external cues and the potential consequences of misplaced trust. In a world where decisions can have cascading effects, understanding the dynamics of sequential decision-making becomes paramount.

\section{Background}
\label{sec:sample1}

Observational learning is crucial in deciphering the nuances of option adoption behaviors. Song and Walden's work on the adoption of peer-to-peer technologies underscores the importance of observational learning \cite{song2003, li2019evaluate}. Walden and Browne's research further elucidates the impact of observational learning on investor behaviors, particularly in their reactions to electronic commerce announcements \cite{walden2008}. Delving into the realm of herd behavior in observational learning, Çelen and Shachar spotlighted the phenomenon of information cascades \cite{celen2004}. This aligns with the broader framework of technology adoption decision-making models, which offer insights into the dynamics of technology adoption and the factors that drive herd behaviors \cite{GU2015471}.

The nuanced interplay between principals and decision-making processes has garnered significant attention in academic circles. Kamenica and Gentzkow's introduction of Bayesian persuasion is a cornerstone in this field, emphasizing optimal signaling in the face of incomplete information in static contexts \cite{kamenica2011}. This model has found resonance across various domains. For instance, implications of this model in the realms of security protocols and voting mechanisms have been explored by Rabinovich et al. \cite{rabinovich2015}. Concurrently, Xu et al. and Goldstein and Leitner delved into its ramifications on investment strategies and market trends \cite{xu2015, goldstein2018}.

As Bayesian persuasion evolved, it paved the way for dynamic models. Foundational works by Ely and Renault et al. set the stage for this progression \cite{ely2017, renault2017}. Cutting-edge research by Celli et al. and Castiglioni et al. tackled the algorithmic intricacies inherent in these dynamic models \cite{celli2020, castiglioni2020b, castiglioni2021}. A fascinating tangent to this is the concept of automated mechanism design, a topic broached by Conitzer and Sandholm \cite{conitzer2002, conitzer2004}. Building on this, Zhang and Conitzer offered a fresh perspective, accentuating the dynamism inherent in principal-agent interactions \cite{zhang2021}.

\section{Principle-WB model}
\label{sec:sample1}
\subsection{Assumptions}
For research in binary adoption processes, it's crucial to grasp the influence of a principle's signal on sequential decision-making. We've built upon the model assumptions set forth by \cite{GU2015471} and \cite{Browne}, introducing additional assumptions to account for the presence of a principle. This enhanced model is denoted as the P-WB model. The following are our assumptions:

\begin{enumerate}
    \item Decision-makers are presented with two distinct options: \( A \) and \( B \).
    \item An unbounded sequence of decision-makers confronts the identical challenge of choosing between \( A \) and \( B \) under comparable situations. These decisions occur sequentially and at distinct time intervals. Furthermore, the decision-making order is exogenous, ensuring that one decision-maker's choice does not influence the sequence or choices of others.
    \item For every decision-maker, the benefits of options \( A \) and \( B \) remain consistent and are unaffected by the timing of the decision-making process.
    \item Each decision maker possesses confidential data regarding the advantages of options \( A \) and \( B \). Specifically, they obtain an objective signal that follows a Gaussian distribution. Given that options \( A \) and \( B \) excel under distinct scenarios, the objective signals they receive originate from separate distributions. If technology \( A \) surpasses \( B \), the associated signal stems from \(N(\mu_A, \sigma^2)\). Conversely, if option \(B\) outperforms option \(A\), the signal stems from \(N(\mu_B, \sigma^2)\). Assuming, for simplicity, that \(\mu_A > \mu_B\), we use \(d = \frac{\mu_A - \mu_B}{\sigma}\) to gauge the disparity between these distributions. For an informed decision, the decision maker must discern which distribution the received objective signal is more likely derived from.
    \item Apart from the objective signal, every decision maker also obtains a signal from a principal. This principal signal carries an inherent bias. To streamline our discussion, we represent this bias as the probability \(P_{\text{bias}}\) associated with the principal opting for option \(A\). If the principal selects \(A\), the principal signal originates from \(N(\mu_A, \sigma_\text{P}^2)\). Conversely, if option \(B\) is chosen, the principal signal originates from \(N(\mu_B, \sigma_\text{P}^2)\).
    \item In addition to the objective signal and the principal signal, decision-makers are aware of the initial conditions and can monitor the choices made by those before them. These initial conditions specify the prior probabilities of option A being superior to B and vice versa, represented as \(P_{\mu_A}\) and \(P_{\mu_B}\) respectively. For simplicity, we set \(P_{\mu_A} = P_{\mu_B} = 0.5\), suggesting an absence of pre-existing bias. While decision-makers can observe others' actions, the underlying motivations remain opaque.
    \item Decision-makers, driven by rationality, ground their choices in available information. The decisions don't diverge from this data-driven approach due to mere personal inclinations. This aligns with the rationality assumption.
\end{enumerate}
Given the assumptions enumerated above, our P-WB model provides a nuanced representation of the decision-making process faced by sequential decision-makers operating under uncertainty. These assumptions encapsulate the necessity of leveraging both objective and subjective signals, as well as previous decisions made by others in determining the optimal choice between options \( A \) and \( B \).

As we proceed, we will delve deeper into how these assumptions play out in real-world scenarios, the potential pitfalls they might present, and the strategies decision-makers can employ to navigate these challenges effectively.

\section{Model Formalization with Principal Influence}

In our refined model, decision-makers receive a continuous objective signal \( s_o \) from a Gaussian distribution, with a mean represented as either \( \mu_A \) or \( \mu_B \). This distinction signifies the superiority of option \( A \) over option \( B \) or vice versa. In addition to this objective signal, decision-makers also receive a principal signal \( s_p \) emanating from a Gaussian distribution with identical mean values, \( \mu_A \) or \( \mu_B \), but different standard deviations, denoted as \( \sigma_p \). This principal influence can potentially modulate the decision-making trajectory.

To make an informed choice, decision-makers need to discern from which Gaussian distribution their signal most likely originates, taking into account both the objective evidence and insights obtained from the choices of their predecessors. The principal's signal, with its unique standard deviation, introduces another layer of complexity to this evaluation.

A decision-maker will be inclined towards option \( A \) if the following relationship is satisfied:
\[
\frac{p(s_o|\mu_A, s_p)}{p(s_o|\mu_B, s_p)} \ge \beta
\]
Using the Gaussian distribution density function, a decision threshold for \( s_o \) is derived as:
\[
r(\beta) = \frac{\sigma^2}{\mu_A - \mu_B} \ln \beta + \frac{\mu_A + \mu_B}{2}
\]
When \( s_o \) is greater than \( r(\beta) \), the decision-maker chooses option \(A\) and When \( s_o \) is less than \( r(\beta) \), the decision-maker chooses option \(B\). Consequently, the following decision rule, termed the G-WB criterion, is established:
\begin{align*}
s_o &\ge r(\beta) &\Rightarrow \text{choosing } A \\
s_o &< r(\beta) &\Rightarrow \text{choosing } B
\end{align*}
This implies:
\begin{itemize}
    \item When \( s_o \) exceeds \(r(\beta)\), the decision-maker chooses \( A \).
    \item Otherwise, the decision-maker selects \( B \).
\end{itemize}

The optimal threshold is selected by balancing the benefits of each technology and the decision maker's information regarding the probability that \( A \) is better than \( B \) and that \( B \) is better than \( A \). In this case, the optimal value of \( \beta \) is
\begin{equation}
\beta = \frac{p_{\mu_B} (\text{Benefit}(b|\mu_B) - \text{Benefit}(a|\mu_B))}{p_{\mu_A} (\text{Benefit}(a|\mu_A) - \text{Benefit}(b|\mu_A))} = \frac{p_{\mu_B}}{p_{\mu_A}} k
\end{equation}
where
\begin{equation}
k = \frac{\text{Benefit}(b|\mu_B) - \text{Benefit}(a|\mu_B)}{\text{Benefit}(a|\mu_A) - \text{Benefit}(b|\mu_A)}
\end{equation}
\( k \) represents the relative difference between the values of the two technologies in different states of the world. The numerator is the difference between the benefit of choosing \( B \) and the benefit of choosing \( A \) in the state where the technology \( B \) is better than \( A \). The denominator is the difference between the benefit of choosing \( A \) and the benefit of choosing \( B \) in the state where the technology \( A \) is better than \( B \). According to the assumption \#3, both the numerator and denominator are positive numbers and \( k \) is a constant.

Incorporating this signal into the Bayesian update:

\begin{align}
p_{\mu_A \cdot (t+1)} &= p(\mu_A | D_t, D_{t-1}, \dots, D_1, s_p) = \frac{ p_{\mu_A} \cdot  p(D_t, D_{t-1}, \dots, D_1 | \mu_A) p(s_p | \mu_A)}{p(D_t, D_{t-1}, \dots, D_1)  p(s_p)} \\
p_{\mu_B \cdot (t+1)} &= p(\mu_B | D_t, D_{t-1}, \dots, D_1, s_p) = \frac{ p_{\mu_B} \cdot p(D_t, D_{t-1}, \dots, D_1 | \mu_B) p(s_p | \mu_B)}{p(D_t, D_{t-1}, \dots, D_1)  p(s_p)}
\end{align}

Where \( p_{\mu_A \cdot (t+1)} = p(\mu_A |s_p, D_t, D_{t-1}, \dots, D_1 ) \) and \( p_{\mu_B \cdot (t+1)} = p(\mu_B |s_p, D_t, D_{t-1}, \dots, D_1 ) \). \( s_p \) is independent with previous observed options \( D_t, D_{t-1}, \dots, D_1 )\) , Thus 

\begin{equation}
\beta_{t+1} = \frac{p_{\mu_B \cdot (t+1)}}{p_{\mu_A\cdot (t+1)}} k 
= \frac{ p(D_t, D_{t-1}, \dots, D_1 | \mu_B) p(s_p | \mu_B)}{ p(D_t, D_{t-1}, \dots, D_1 | \mu_A) p(s_p | \mu_A)} \left[\frac{p_{\mu_B}}{p_{\mu_A}} k\right]
\end{equation}  
and define \( A_t \) as \( A_t = \{D_{t-1}, D_{t-2}, \dots, D_1\} \), the set of all prior decisions for all \( t > 1 \), then
\begin{align}
p(D_t, D_{t-1}, \dots, D_1 | \mu_A) &= p(D_t | A_t, \mu_A) p(D_{t-1}, \dots, D_1 | \mu_A) \\
p(D_t, D_{t-1}, \dots, D_1 | \mu_B) &= p(D_t | A_t, \mu_B) p(D_{t-1}, \dots, D_1 | \mu_B)
\end{align}
Thus, we obtain the iterative formula:
\begin{equation}
\beta_{t+1} = \frac{p(D_t | A_t, \mu_B) p(D_{t-1}, \dots, D_1 | \mu_B)}{p(D_t | A_t, \mu_A) p(D_{t-1}, \dots, D_1 | \mu_A)} \left[\frac{ p(s_p | \mu_B) p_{\mu_B}}{ p(s_p | \mu_A) p_{\mu_A}} k \right] = \frac{p(D_t | A_t, \mu_B) p(s_p | \mu_B)}{p(D_t | A_t, \mu_A) p(s_p | \mu_A)} \beta_t
\label{principal_decision}
\end{equation}
In this manner, we can obtain the threshold of the t-th decision maker and the probability of the t-th decision maker’s choice after observing the decisions made by the previous decision-makers and the signal from the principle.

\section{Data and Results}
\subsection{Decision Dynamics Absent of Principal Influence}
\label{Decisionnoprincipal}

Within the realm of social systems, decision-making dynamics often manifest complex patterns. A notable example is the 'herd behavior', where, over time, a majority of participants converge towards a single choice. It's noteworthy that subsequent decision-makers often base their choices solely on the observed decision, without access to the underlying rationale that informed it.

For a clearer understanding of this dynamic, we set our model parameters as: \( \mu_A = 1 \), \( \mu_B = 0 \), \( \sigma = 1 \), and \( k = 1 \). We assumed that the objective signal is uniformly drawn from \(N(\mu_A, \sigma^2) \). The outcomes from this configuration align with the foundational works of \cite{GU2015471}. Specifically, there was a consistent rise in the proportion of participants opting for choice \(A\). This aligns with our model's design, where the decision threshold, denoted by \(r(\beta) \), progressively shifts towards \(\mu_B\), diminishing the influence of the objective signal in subsequent decisions.

\begin{figure}[h]
    \centering
    \includegraphics[width=1\textwidth]{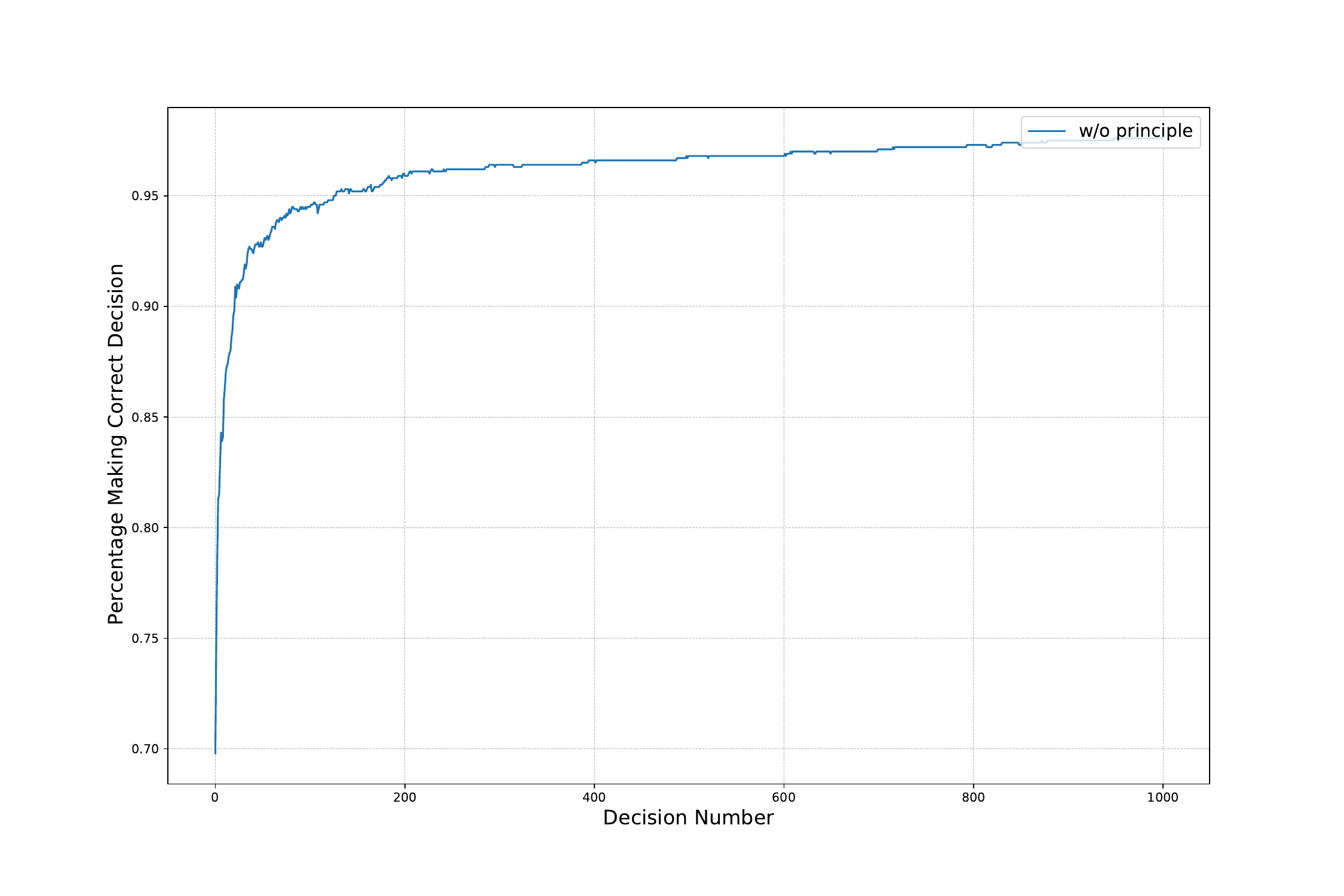}
    \caption{Convergence trends in the WB model without principal influence.}
    \label{fig:Percentage_NoPrinciple}
\end{figure}

Fig.\ref{fig:Percentage_NoPrinciple} vividly portrays the proportion of decision-makers selecting the correct choice, as a function of decision count. This pattern of convergence towards the correct choice, which intensifies with an increasing number of decision-makers, echoes the findings of \cite{Browne,GU2015471}.

\subsection{Decision Dynamics Influenced by Unwaveringly Trusted Principals}
\label{Decisiontrustedprincipal}
Introducing principals into the decision-making framework adds layers of complexity. Notably, in our model, principals aren't necessarily reliable sources of the correct signal, here denoted as option \(A\). We utilize \(p_b\) to represent the likelihood that a principal's advice originates from the correct signal distribution \(N(\mu_A, \sigma^2)\). If not, the advice aligns with the incorrect signal distribution \(N(\mu_B, \sigma^2)\).

\begin{figure}[h]
    \centering
    \includegraphics[width=1\textwidth]{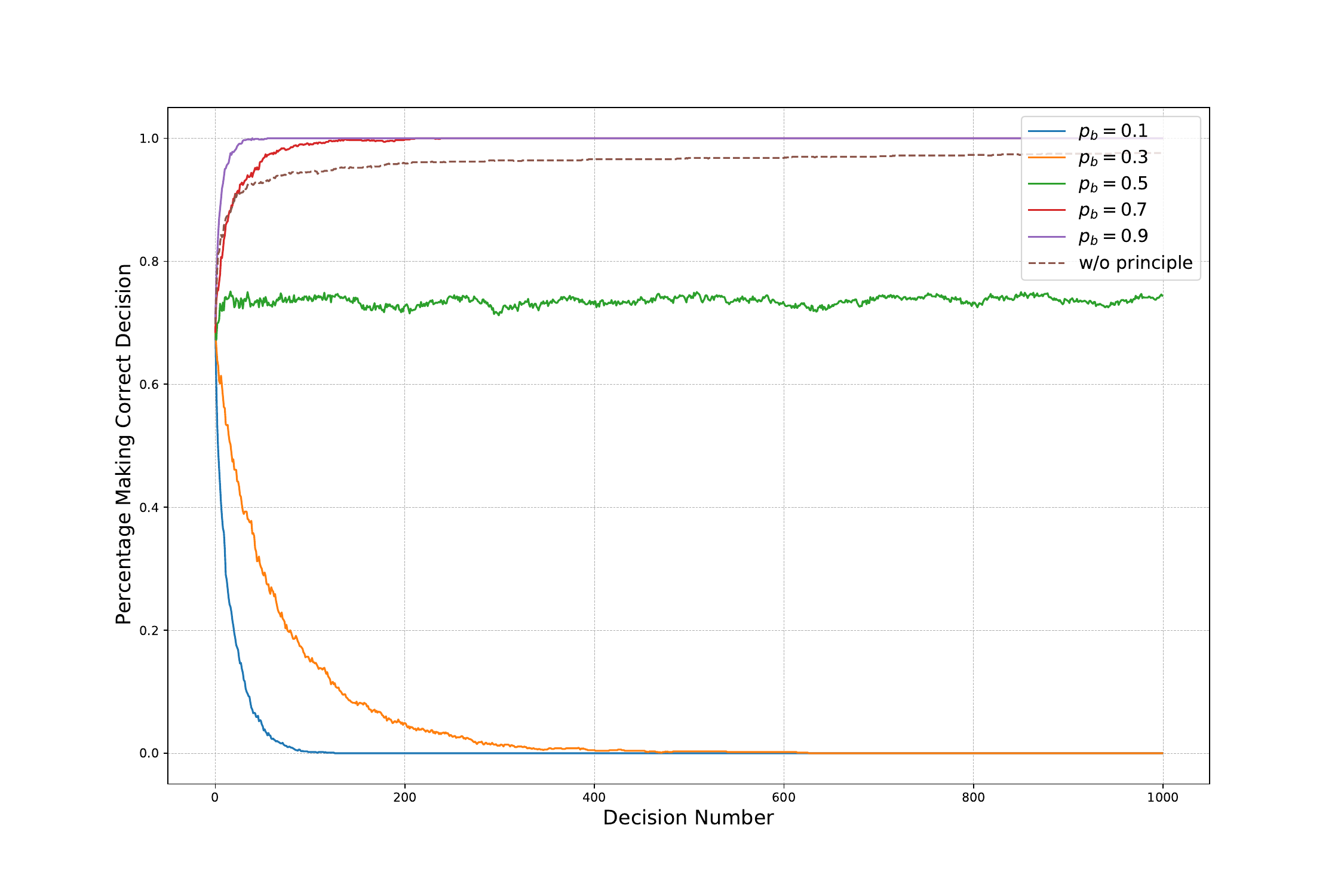}
    \caption{Dynamics under the influence of the P-WB model.}
    \label{fig:Percentage_Principle}
\end{figure}

Fig.\ref{fig:Percentage_Principle} offers a comparative study across various \(p_b\) values. The influence of the principal is palpable, as the decision-makers' convergence trajectory seems significantly influenced by the principal's signal accuracy. This influence remains pronounced even when private signals consistently arise from \(N(\mu_A, \sigma^2)\). For a comprehensive comparison, the principal-free scenario from Sec.\ref{Decisionnoprincipal} is also depicted.

A thorough assessment reveals several insights:

\begin{enumerate}
    \item Principals with a high accuracy rate expedite convergence towards the optimal decision.
    \item In contrast, principals prone to errors can misguide decisions towards less optimal outcomes.
    \item When the principal's accuracy surpasses a critical 50\% threshold, the initial convergence might be gradual, but eventually surpasses the baseline, highlighting the principal's dominant influence.
    \item Interestingly, at \(p_b = 0.5\), decision patterns stabilize, showing no evident convergence.
\end{enumerate}

\subsection{Decision Dynamics Influenced by Conditionally Trusted Principals}
In Sec.\ref{Decisiontrustedprincipal}, we considered a principal that consistently garners trust from the decision-maker, illustrating how decision convergence is largely governed by the principal's influence. Here, we delve into a more intricate scenario: the principal isn't invariably trusted, irrespective of its signal's accuracy. In this context, \(p_t\) denotes the likelihood of the decision-maker trusting the principal's signal. When trusted, the decision boundary is adjusted as per Eq.\ref{principal_decision}. If not, the decision-maker disregards the principal's signal, updating the decision boundary based on \cite{Browne}:
\begin{equation}
\beta_{t+1} = \frac{p(D_t | A_t, \mu_B)}{p(D_t | A_t, \mu_A)} \beta_t
\label{noprincipal_decision}
\end{equation}

\begin{figure}[h]
    \centering
    \includegraphics[width=1\textwidth]{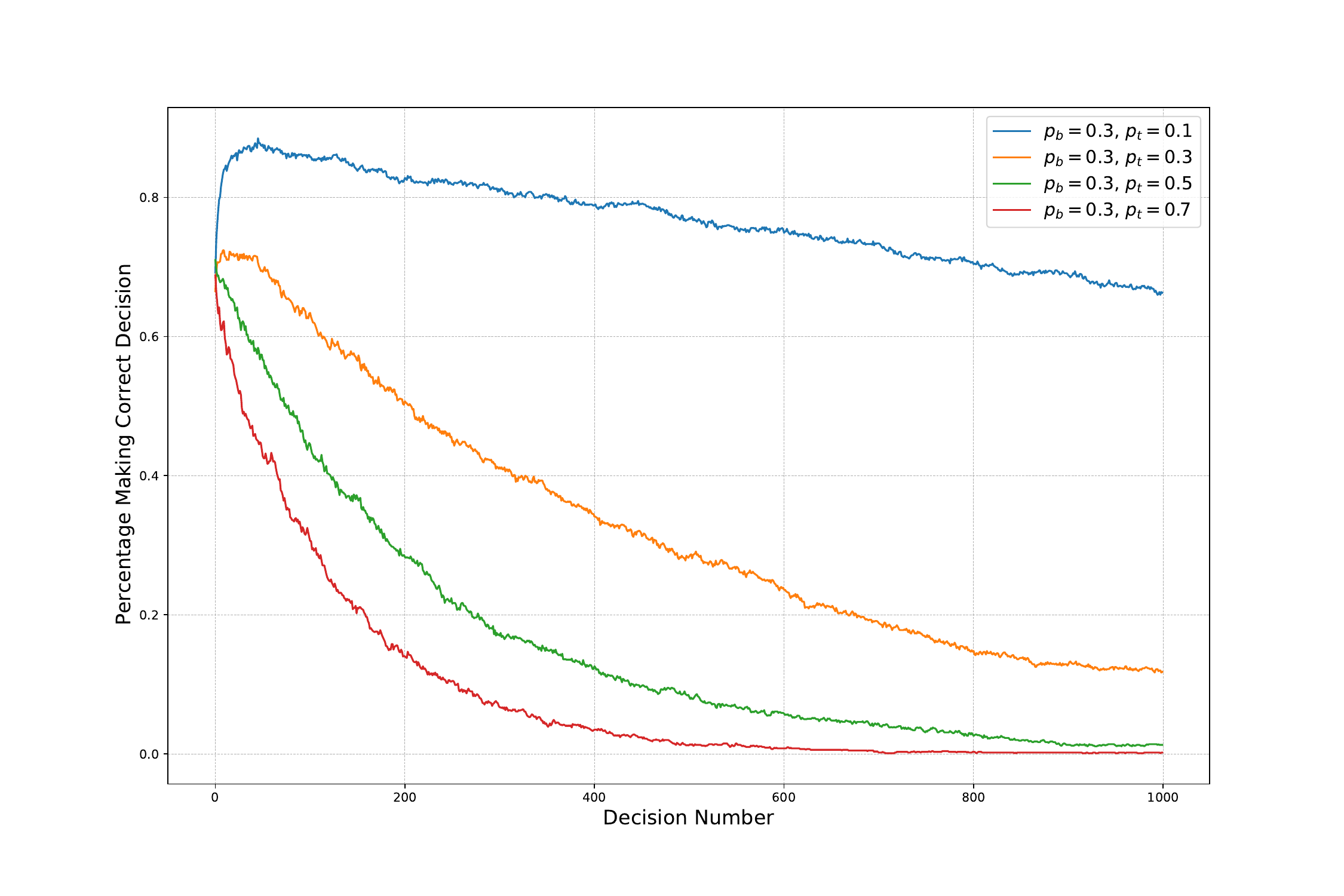}
    \caption{Decision dynamics under varying trust levels.}
    \label{fig:Percentage_Principle_trustness}
\end{figure}

Our analysis first contrasts results across different \(p_t\) values, given \(p_b = 0.3\). This setting implies the principal's guidance is accurate 30\% of the time, and even so, is trusted only 10\% of the time by decision-makers. The outcomes, as visualized in Fig.\ref{fig:Percentage_Principle_trustness}, are compelling. Initially, there's a rise in the percentage of correct decisions, hinting at decision-makers benefiting from prior accurate decisions. Yet, as decisions accumulate, this trend inverts, leading to a decline in decision accuracy. This unexpected result emphasizes a crucial observation: even a predominantly misleading principal, if occasionally trusted, can profoundly alter decision-making trajectories. This effect can be traced back to prior studies indicating that accurate guidance from a trusted principal reinforces convergence towards the correct decision. However, in this context, as more decisions are swayed by the principal's erroneous guidance, the overall decision accuracy wanes.

This finding aligns with our study's overarching theme, highlighting the significant influence of external entities, like principals, on sequential decision-making. Even when such influences are intermittent and potentially unreliable, they can shape collective decision-making patterns, emphasizing the importance of discernment in decision-making trust allocation.

\begin{figure}[h]
    \centering
    \includegraphics[width=1\textwidth]{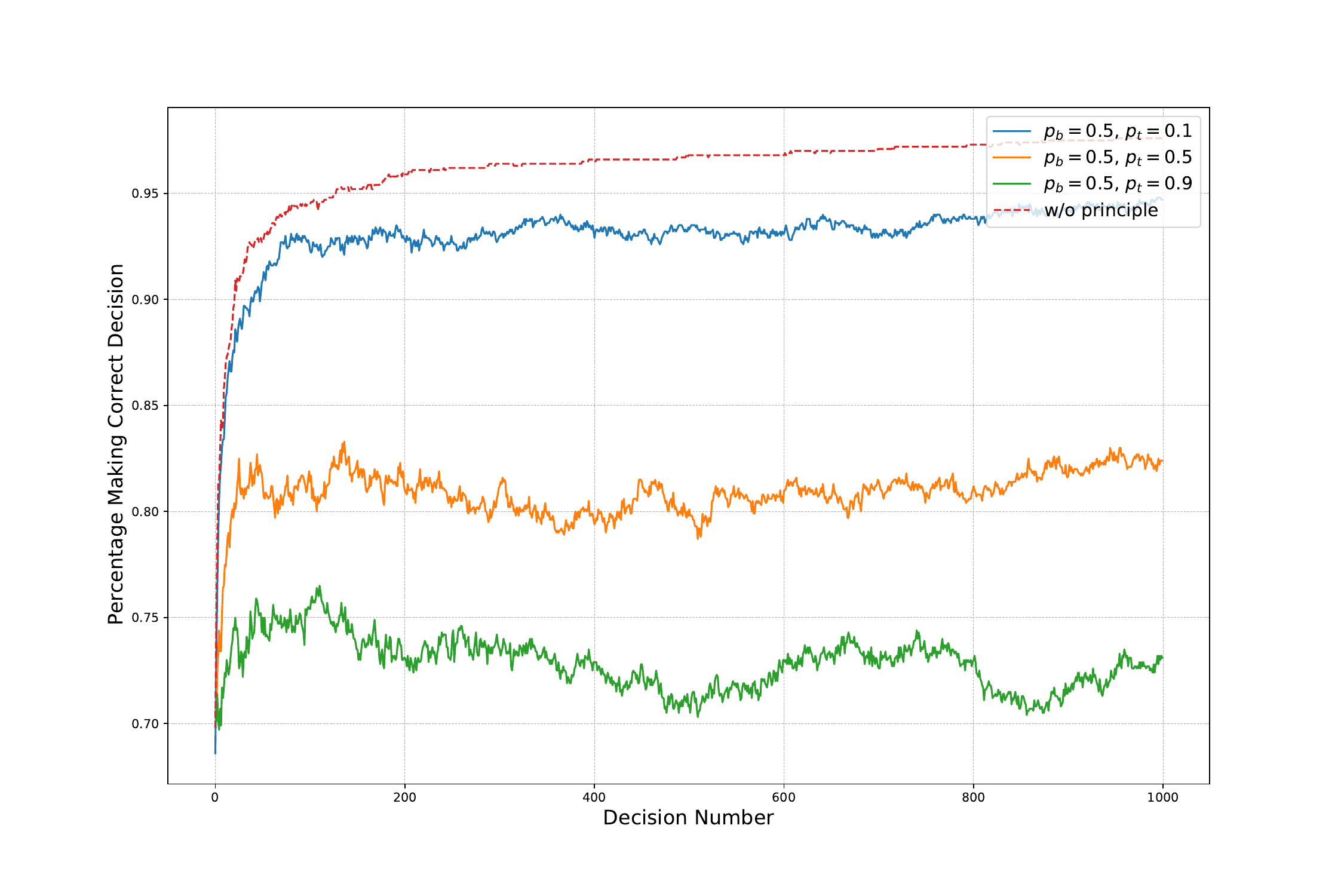}
    \caption{Decision dynamics under varying trust levels with a 50\% bias parameter.}
    \label{fig:Percentage_Principle_trustness_halfbias}
\end{figure}

In a subsequent phase of our analysis, we set \(p_b\) to 0.5, suggesting that the principal's guidance is accurate half the time, making it statistically random. Fig.\ref{fig:Percentage_Principle_trustness_halfbias} charts decision-making accuracy against diverse \(p_t\) values. The results reveal an unexpected inverse correlation: as trust in the principal grows, decision accuracy recedes. This suggests that increased trust in a statistically random principal can inadvertently misguide decision-makers.

While one might anticipate a neutral impact from a random principal on decision-making, the data indicates otherwise. Entrusting an external source, even one offering statistically random guidance, can introduce variability into the decision-making process. This variability, when compounded over successive decisions, can lead to deviations from optimal decision pathways.

This phase of the analysis underscores the intricate interplay between trust and decision-making accuracy. Even without a discernible bias in external guidance, the mere act of entrusting it can introduce uncertainties, affecting the broader decision-making framework. This emphasizes the importance of prudence in trust allocation, especially when the reliability of external guidance is ambiguous or unverified.

\section{Summary}

Our comprehensive analysis delved into the intricate dynamics of decision-making, especially within social systems. We explored the phenomenon of 'herd behavior', as described in previous works, where participants tend to converge towards a singular choice over time. This behavior was characterized by subsequent decision-makers basing their choices solely on the observed decision, without access to the underlying rationale.

In scenarios devoid of principal influence, our model's outcomes aligned with foundational works, showcasing a consistent preference for optimal option. This was attributed to the model's design, where the decision threshold progressively shifts, reducing the influence of the objective signal in subsequent decisions.

However, the introduction of principals, entities that may or may not provide reliable signals, added layers of complexity. The decision-making trajectory was significantly influenced by the principal's signal accuracy. Even when private signals were consistently accurate, the principal's influence remained dominant. 

In scenarios where the principal wasn't consistently trusted, even occasional trust in a predominantly misleading principal could profoundly alter decision-making trajectories. This emphasized the significant influence of external entities on sequential decision-making and the importance of discernment in decision-making trust allocation.

Interestingly, when the principal's guidance was statistically random, an inverse relationship between trust in the principal and decision accuracy emerged. This highlighted the potential pitfalls of placing undue trust in external sources, especially when their reliability is uncertain.

In essence, our study underscores the nuanced interplay between trust, external influences, and decision-making accuracy. It emphasizes the importance of prudence in trust allocation and the potential consequences of misplaced trust, especially in complex decision-making landscapes.
\appendix

 \bibliographystyle{elsarticle-num} 
 \bibliography{cas-refs}

\begin{thebibliography}{10}
\expandafter\ifx\csname url\endcsname\relax
  \def\url#1{\texttt{#1}}\fi
\expandafter\ifx\csname urlprefix\endcsname\relax\def\urlprefix{URL }\fi
\expandafter\ifx\csname href\endcsname\relax
  \def\href#1#2{#2} \def\path#1{#1}\fi

\bibitem{song2003}
J.~Song, E.~Walden, Consumer behavior in the adoption of peer-to-peer technologies: an empirical examination of information cascades and network externalities (2003).

\bibitem{li2019evaluate}
L.~Li, Z.~He, X.~Zhou, D.~Yu, How to evaluate the next system: Automatic dialogue evaluation from the perspective of continual learning (2019).
\newblock \href {http://arxiv.org/abs/1912.04664} {\path{arXiv:1912.04664}}.

\bibitem{walden2008}
E.~Walden, G.~Browne, Rational fads in investor reactions to electronic commerce announcements: an explanation of the internet bubble, Electron. Commer. Res. Appl. 7 (2008) 44--54.

\bibitem{celen2004}
B.~Çelen, K.~Shachar, Observational learning under imperfect information, Games Econ. Behav. 47 (2004) 72--86.

\bibitem{GU2015471}
J.~Gu, L.~Li, Z.~Xu, H.~Fujita, \href{https://www.sciencedirect.com/science/article/pii/S0950705115003251}{Construction of a technology adoption decision-making model and its extension to understanding herd behavior}, Knowledge-Based Systems 89 (2015) 471--486.
\newblock \href {https://doi.org/https://doi.org/10.1016/j.knosys.2015.08.014} {\path{doi:https://doi.org/10.1016/j.knosys.2015.08.014}}.
\newline\urlprefix\url{https://www.sciencedirect.com/science/article/pii/S0950705115003251}

\bibitem{kamenica2011}
E.~Kamenica, M.~Gentzkow, Bayesian persuasion, American Economic Review 101~(6) (2011) 2590--2615.

\bibitem{rabinovich2015}
Z.~Rabinovich, et~al., Information disclosure as a means to security (2015) 645--653.

\bibitem{xu2015}
H.~Xu, et~al., Exploring information asymmetry in two-stage security games, Manage. Inform. Syst. Quart. 33 (2015) 23--48.

\bibitem{goldstein2018}
I.~Goldstein, Y.~Leitner, Stress tests and information disclosure, Journal of Economic Theory 177 (2018) 34--69.

\bibitem{ely2017}
J.~Ely, Beeps, American Economic Review 107~(1) (2017) 31--53.

\bibitem{renault2017}
J.~Renault, E.~Solan, N.~Vieille, Optimal dynamic information provision, Games and Economic Behavior 104 (2017) 329--349.

\bibitem{celli2020}
A.~Celli, et~al., Private bayesian persuasion with sequential games, in: Proceedings of the AAAI Conference on Artificial Intelligence, Vol.~34, 2020, pp. 1886--1893.

\bibitem{castiglioni2020b}
M.~Castiglioni, et~al., Online bayesian persuasion, Advances in Neural Information Processing Systems 33 (2020) 16188--16198.

\bibitem{castiglioni2021}
M.~Castiglioni, et~al., Multi-receiver online bayesian persuasion, in: Proceedings of 38th International Conference on Machine Learning, 2021, pp. 1314--1323.

\bibitem{conitzer2002}
V.~Conitzer, T.~Sandholm, Complexity of mechanism design, in: Proceedings of the 18th Conference in Uncertainty in Artificial Intelligence, 2002, pp. 103--110.

\bibitem{conitzer2004}
V.~Conitzer, T.~Sandholm, Self-interested automated mechanism design and implications for optimal combinatorial auctions, in: Proceedings of the 5th ACM Conference on Electronic Commerce, 2004, pp. 132--141.

\bibitem{zhang2021}
H.~Zhang, V.~Conitzer, Automated dynamic mechanism design, Advances in Neural Information Processing Systems 34 (2021).

\bibitem{Browne}
E.~Walden, G.~Browne, Sequential adoption theory: A theory for understanding herding behavior in early adoption of novel technologies, J. AIS 10 (01 2009).
\newblock \href {https://doi.org/10.17705/1jais.00181} {\path{doi:10.17705/1jais.00181}}.

\end{thebibliography}





% \end{thebibliography}
\end{document}